\documentclass[10pt,twocolumn,letterpaper]{article}

\usepackage{ijcb}
\usepackage{times}
\usepackage{epsfig}
\usepackage{graphicx}
\usepackage{amsmath}
\usepackage{amssymb}
\usepackage{caption}
\usepackage{subcaption}
\usepackage{gensymb}
\usepackage{booktabs, multicol, multirow}
\usepackage{array}
\usepackage{color}
\usepackage{cite}
\usepackage{balance}
\usepackage{setspace}
\usepackage{csquotes}

\ijcbfinalcopy 

\begin{document}

\title{Pose Attention-Guided Profile-to-Frontal Face Recognition}

\author{ Moktari Mostofa, \ Mohammad Saeed Ebrahimi Saadabadi, \ Sahar Rahimi Malakshan, and \ Nasser M. Nasrabadi\\
West Virginia University\\
{\tt \footnotesize \{mm0251,me00018,sr00033\}@mix.wvu.edu,  nasser.nasrabadi@mail.wvu.edu}}


\maketitle
\thispagestyle{empty}

\begin{abstract}
In recent years, face recognition systems have achieved exceptional success due to promising advances in deep learning architectures. However, they still fail to achieve expected accuracy when matching profile images against a gallery of frontal images. Current approaches either perform pose normalization (i.e., frontalization) or disentangle pose information for face recognition. We instead propose a new approach to utilize pose as an auxiliary information via an attention mechanism. In this paper, we hypothesize that pose attended information using an attention mechanism can guide contextual and distinctive feature extraction from profile faces, which further benefits a better representation learning in an embedded domain. To achieve this, first, we design a unified coupled profile-to-frontal face recognition network. It learns the mapping from faces to a compact embedding subspace via a class-specific contrastive loss. Second, we develop a novel pose attention block (PAB) to specially guide the pose-agnostic feature extraction from profile faces. To be more specific, PAB is designed to explicitly help the network to focus on important features along both \enquote{channel} and \enquote{spatial} dimension while learning discriminative yet pose-invariant features in an embedding subspace. To validate the effectiveness of our proposed method, we conduct experiments on both controlled and in-the-wild benchmarks including Multi-PIE, CFP, IJB-C, and show superiority over the state-of-the-arts. 
\end{abstract}
\vspace{-6mm}
\section{Introduction}

The advent of deep convolutional neural networks (CNNs) has led to promising achievement in unconstrained face recognition and verification techniques \cite{sun2014deep, taigman2014deepface}. It has even surpassed the human performance on several benchmark datasets \cite{schroff2015facenet}. However, a challenge that still remains to be solved, is that of extreme pose variations in profile faces. It degrade frontal-to-profile face verification accuracy by more than 10\% compared to frontal-to-frontal matching accuracy \cite{sengupta2016frontal}. The most prominent factors contributing to this performance degradation can be classified into three categories:
\vspace{-2mm}
\begin{figure}[t]
\centering
\includegraphics[width=5.5cm]{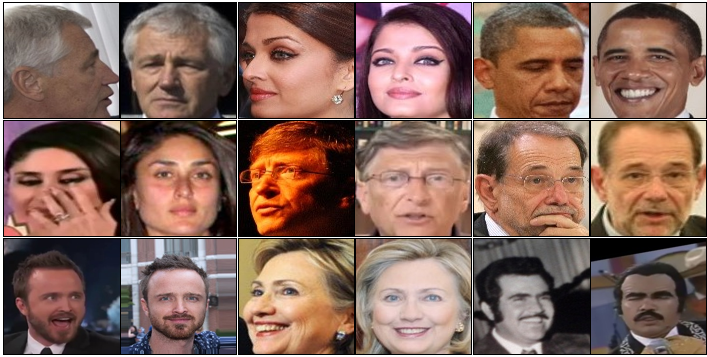}
\caption{Frontal and Profile faces in the IJB-C Dataset under full pose variation, expression, and different imaging conditions.}\label{fig:profile_frontal_images}
\vspace{-5mm}
\end{figure}
\begin{itemize}
    \item Facial appearance distortion: In comparison to controlled environment, real world profile faces have different imaging conditions besides pose such as expression, occlusion and illumination variations as shown in Figure 1. These variations cause substantial changes in facial appearance, which indicates loss of consistent information useful for face recognition. 
 \vspace{-2mm}
    \item Missing semantic consistency: When a face view is changed from frontal to profile, the position and shape of facial texture varies nonlinearly. It inevitably introduces loss of semantic correspondence in 2D images along with confusion in interpersonal texture differences \cite{ding2016comprehensive}. In consequence, extracted features from two images at different poses are no longer similar and cannot provide high matching accuracy as expected in conventional face recognition methods.
    \vspace{-2mm}
    \item Imbalance pose distributions: Deep learning based face recognition algorithms extensively rely on very large datasets, which usually suffer from uneven pose distributions. This data imbalance continues to force the model to lean towards frontal images than profile face image of a person. It results in poor matching accuracy during frontal-to-profile verification. In contrast, human can easily identify faces with extreme pose variations without significant drop in accuracy.
\end{itemize}

\vspace{-0.5mm}
To address this performance gap between human and automatic models, traditional methods apply several local descriptors such as Gabor \cite{daugman1985uncertainty}, Haar \cite{viola2001rapid}, and LBP \cite{ahonen2006face} to measure local distortions and then adopt metric learning techniques \cite{chen2013blessing, weinberger2009distance}. On the other hand, another research community emphasizes on frontal view synthesis across poses. They utilize 3D geometrical transformations\cite{hassner2015effective, taigman2014deepface, zhu2015high} to reduce pose variations. Moreover, multiple novel architectures have been proposed\cite{zhao2018towards, qian2019unsupervised, tran2017disentangled} for face normalization, which aligns faces to a canonical pose. Although, they show impressive performance at normalizing small pose faces, their accuracy drops severely under extreme pose conditions. To handle this problem, some researchers opt on learning pose-robust features \cite{cao2018pose,shi2020towards,yin2017multi}
for multi-view face images. Among them, Cao et al. \cite{cao2018pose} propose a lightweight DREAM block to perform \enquote{frontalization} in feature space, while others explore multi-task learning to perform pose-invariant face recognition (PIFR)\cite{ding2015multi,yin2017multi}.

In this paper, we introduce a novel method to learn discriminative pose-invariant representation in a deep feature embedding subspace without performing profile face normalization (frontalization) or learning disentangled features. Instead, we explicitly deal with the pose variability by incorporating it as an \enquote{auxiliary information} via an attention mechanism to the feature extraction network. We hypothesize that learning with pose information allows for better generalization of the primary task by assisting it to focus on the current context and ignore unnecessary information. To this end, we first develop a deep coupled profile to frontal network using the contrastive loss, which is able to learn to map faces into a common compact 512-dimensional latent embedding subspace. 
Second, to incorporate pose as an auxiliary signal, we propose an easy-to-implement pose attention block (PAB), which automatically infers significant features from profile faces along channel and spatial axes in deeper layers of the network. In other words, PAB is designed to empirically guide to learn discriminative and pose-invariant features in an embedding subspace. Moreover, we also investigate the capability of these learned embedding features via a generative adversarial network (GAN) to synthesize a canonical (frontal) view.  In a summary, this paper offers the following contributions:

\begin{itemize}
\vspace{-2mm}
\item  A novel coupled profile to frontal PIFR model utilizing pose as an auxiliary information (i.e., pose attention) is developed.
\vspace{-1mm}
\item A pose attention block (PAB) using a pretrained pose-estimation network is proposed to guide a discriminative and pose-invariant feature learning framework in an embedding subspace. 
\vspace{-1mm}
\item Extensive experiments on different benchmark datasets and comparison to other state-of-the-art methods have been performed to validate the effectiveness of our proposed method.
\vspace{-1mm}
\item Capability of the embedding features learned in our proposed network is explored for frontal face synthesis via a GAN model, which indicates its usefulness in different face analysis tasks apart from face recognition.
\end{itemize}

\section{Related Work}
\vspace{-1.8mm}
\subsection{Face Frontalization}
\vspace{-2.1mm}
Face frontalization has become an extremely challenging task due to the self-occlusion that exists in 2D projections of the input face with large pose variations. To address this problem, traditional methods use 3D based models \cite{hassner2015effective, li2012morphable, zhu2015high}, statistical approaches \cite{sagonas2015robust}, and deep learning based methods \cite{kan2014stacked, yang2015weakly, yim2015rotating, zhu2014multi} for face frontalization. Hassner et al. \cite{hassner2015effective} used a 3D face model to generate frontal shape of all input faces. Although it is proved to be efficient in face frontalization task, it cannot achieve expected accuracy for profile and near profile faces, specifically faces with yaw angle greater than $60\degree$. A statistical model is proposed in\cite{sagonas2015robust}, which solves a constrained low-rank minimization problem to jointly perform frontal view reconstruction and landmark detection. Recently, deep learning based methods have shown outstanding performances in frontal face synthesis. In \cite{yang2015weakly}, a recurrent transform unit is proposed to reconstruct discrete 3D views. Yim et al. \cite{yim2015rotating} applied a concatenated network structure to rotate a non-frontal face, where they regularize the output by image level reconstruction loss. With the emergence of GAN, researchers have concentrated more on GAN-based methods, which has advanced the performance of face frontalization methods. However, face frontalization is considered as an image-level pose-invariant representation, which can improve PIFR performance mostly for face images at near frontal or half profile.


\vspace{-2.4mm}

\begin{figure*}[t]
\centering
    \includegraphics[width=0.64\textwidth,height=7.7cm]{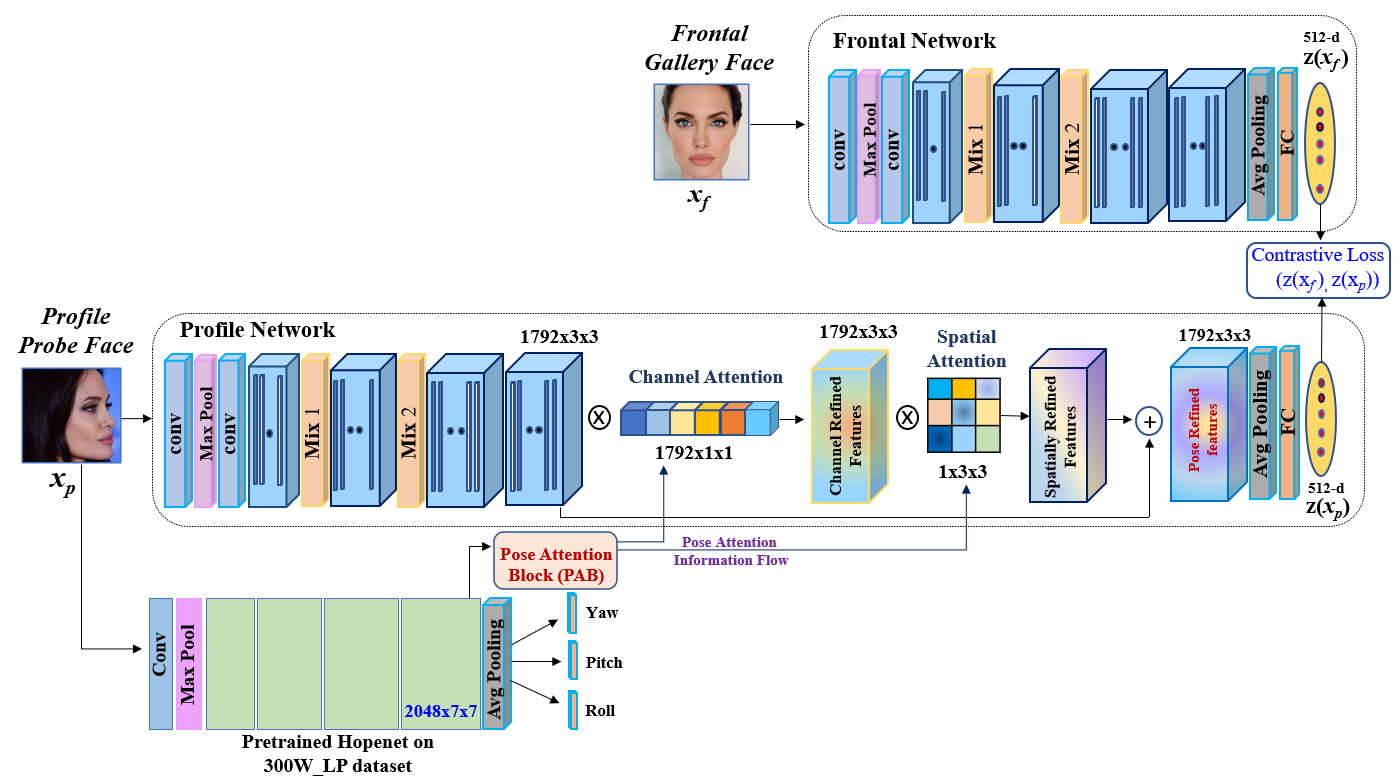}
    \caption{Block diagram of our proposed deep coupled profile-to-frontal PIFR network}
    \label{}
\end{figure*}

\subsection{Pose Invariant Representation Learning}
\vspace{-2mm}
 Pose-invariant feature representation has been recently used as a mainstay of many face recognition tasks. Earlier works apply canonical correlation analysis (CCA)\cite{hardoon2004canonical} to analyze the shared characteristics among pose-invariant samples. Recent deep learning based approaches focus on several aspects while training a network. To name, in \cite{zhu2014multi}, a deep neural network is trained to separate face identity from viewpoints. Kan et al. \cite{kan2016multi} propose feature pooling across different poses to allow a single network structure for inputs at multiple pose views. To disentangle poses in feature representation, several methods \cite{yin2017multi, zhu2013deep} carefully factorize out the non-identity part. Authors in \cite{chen2016unconstrained, Masi_pose_aware_2016} mostly consider fusing information at the feature level or distance metric level. On the other hand, Cao et al.\cite{cao2018pose} propose a pose discrepancy corrector module. Recently many researchers \cite{huang2021attention, tsai2021pam} followed them to empirically perform frontalization in feature space. Contrary to these approaches, we mostly concentrate to utilize pose as side information via an attention mechanism and guide the network to learn discriminative, and pose-invariant features in an embedding subspace. 
 \vspace{-0.25cm}
\section{Proposed Method}
\vspace{-1.3mm}
Here, we describe our proposed method which offers a new perspective of learning pose-invariant feature representation via incorporating pose specific auxiliary information into our deep profile to frontal face verification network. Inspired by the success of FaceNet face recognition system\cite{schroff2015facenet}, we develop a deep coupled framework as shown in Figure 2. It learns mapping from both frontal and profile faces to a compact feature embedding subspace. Since profile faces have large pose variations, we use this angular knowledge to explore how auxiliary pose information can improve the embedding feature representation for profile faces. To implement this perspective, we propose the PAB module, which sequentially refine features in both channel and spatial dimension. In this section, first we discuss the implementation technique of our PAB module, and then we detail how we integrate the auxiliary pose information to our deep coupled network via a channel and spatial attention mechanism.

\subsection{Pose Attention Block (PAB)}
We adopt a robust pose estimation network, i.e., Hopenet \cite{ruiz2018fine}, which has been trained on a large synthetically expanded dataset 300W-LP\cite{zhu2016face}. Hopenet uses ResNet50 as the backbone of their architecture and adds three fully connected layers to predict the intrinsic Euler angles (yaw, pitch and roll) directly from input off-angle face images as illustrated in Figure 2. To implement our proposed PAB module, we do not use these angles, instead we take the feature map of size $2048\times7\times7$ from the last convolutional layer of Hopenet. It already provides us with more complex abstract features such as overall shapes, pose and texture of the input face.

Our proposed PAB module consists of two sequential attention modules: adaptive channel attention module (ACAM), and spatial attention module (SpAM), which aggregate pose information through inferring a 1D channel attention map and a 2D spatial attention map. Figure 3 illustrates the framework of the proposed PAB, which is integrated with our deep coupled learning framework in Figure 2. We now discuss each component in detail.


\vspace{-5mm}

\begin{figure}[t]
    \centering
\begin{subfigure}[b]{0.65\textwidth}
        \includegraphics[width=0.80\textwidth]{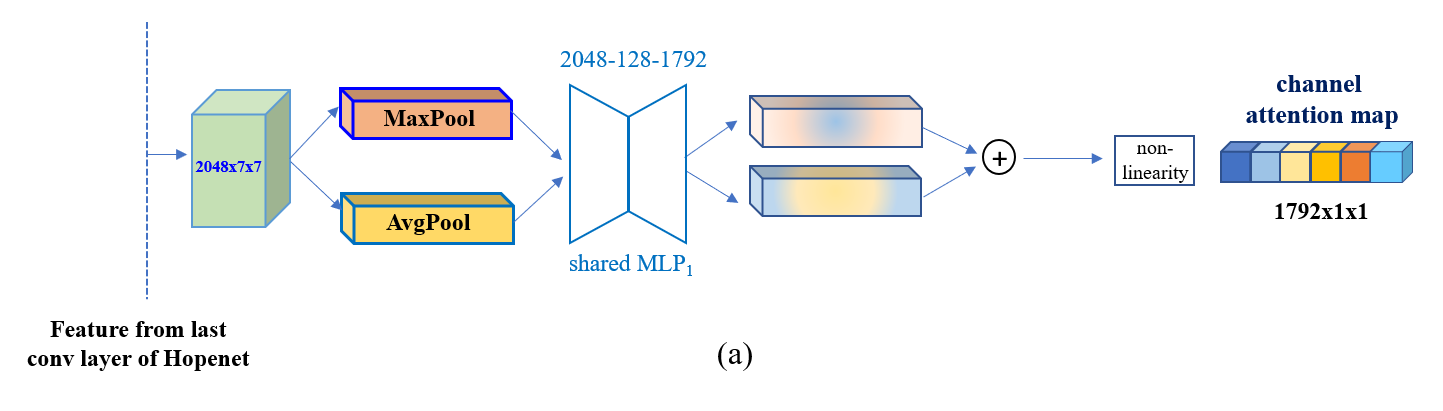}
         \label{fig:y equals x}
     \end{subfigure}
     \hfill
\begin{subfigure}[b]{0.65\textwidth}
         \includegraphics[width=0.70\textwidth]{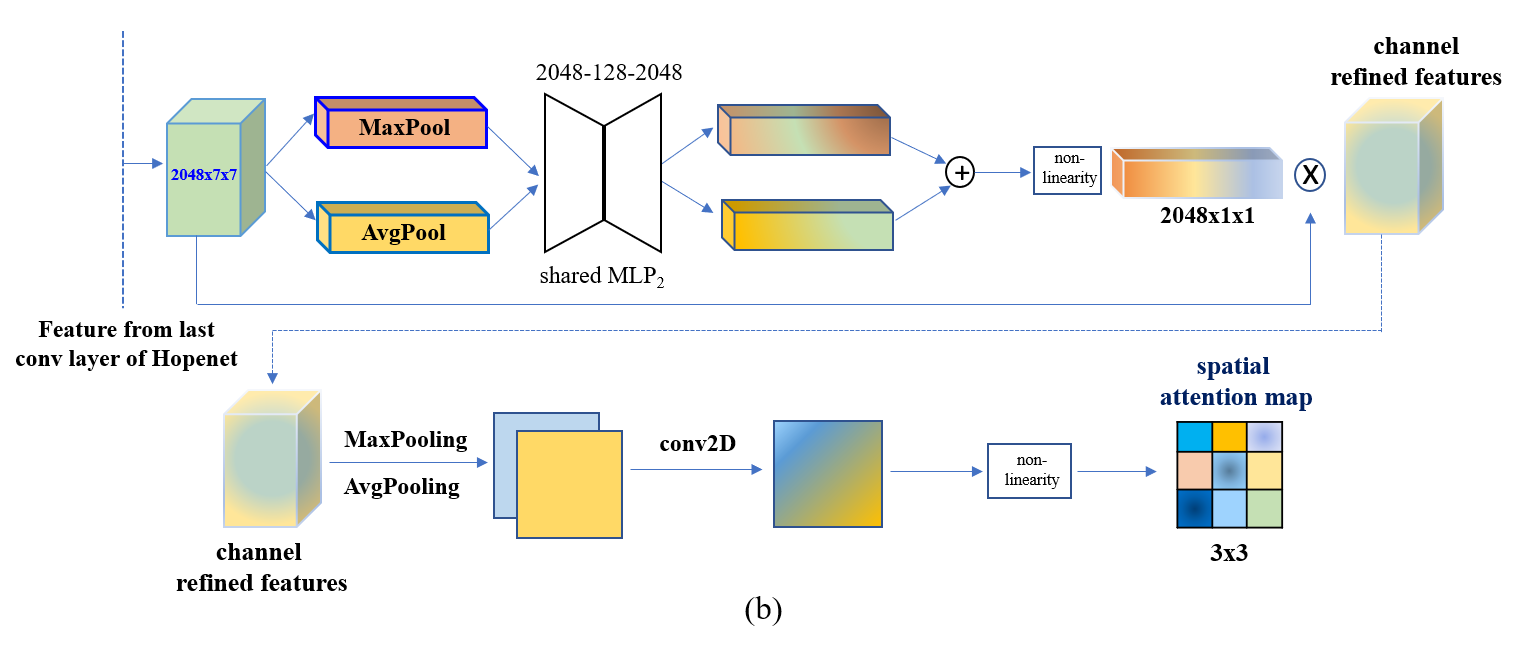}
         \label{fig:three sin x}
     \end{subfigure}
     \caption{Block diagram of our PAB (a) Adaptive Channel attention module (ACAM) (b) Spatial Attention Module (SpAM)}
\vspace{-0.4cm}
 \end{figure}

\subsubsection{Adaptive Channel Attention Module (ACAM)}
\vspace{-1.8mm}
Given an input feature map, $x$ \textbf{$\epsilon$} $\mathbb{R}^{C \times H\times W}$ of size $2,048\times7\times7$, ACAM applies an average-pooling and a max-pooling operation like in CBAM \cite{woo2018cbam}
to learn inter-channel dependencies and generates two different spatial context descriptors: ${x^{c}_{avg}}$, and ${x^{c}_{max}}$, respectively. To integrate spatial information, they are forwarded to a shared multilayer perceptron ($MLP_{1}$), which is typically a two layer fully connected network as shown in Figure 3(a). Since our ultimate goal is to distill features for the face recognition task along the channel dimension in order to highlight \textit {\enquote{which of the feature maps are relevant to shape}}, we set hidden-layer size to $128$ and output in a way that it can generate a 1D channel attention map of $1,792$ consistent with the feature map depth size in profile network. After that, we use element-wise summation to merge feature vectors obtained from the shared $MLP_{1}$ network. In a summary, the channel attention is computed as follows:
\vspace{-2.7mm}

\begin{equation}
M_c(x)=\sigma(MLP_{1}({x^{c}_{avg}})+MLP_{1}({x^{c}_{max}})),
\end{equation}where $\sigma$ is the non-linear activation function. 

\vspace{-0.25cm}
\subsubsection{Spatial Attention Module (SpAM)}

Following CBAM \cite{woo2018cbam}, we design our spatial attention module (SpAM) to focus on the informative region in the spatial domain. In our SpAM, we compute a 2D spatial attention map of size $3\times3$ to be consistent with the feature map size of our FR network as follows:
\vspace{-0.20cm}
\begin{equation}
F_c(x)=(\sigma(MLP_{2}({x^{c}_{avg}})+MLP_{2}({x^{c}_{max}})))\otimes x,
\end{equation}
\vspace{-0.28cm}
\begin{equation}
M_s(x)=\sigma(conv2D[{F^{s}_c(x)_{avg}},{F^{s}_c(x)_{max}}]),
\end{equation} where $F{_c}(x)$ denotes channel refined features in Figure 3(b), which is obtained using another shared $MLP{_2}$. conv2D refers to the convolution on the concatenation of the average pooled feature map, $F^{s}_c(x)_{avg}$, and max pooled feature map, $F^{s}_c(x)_{max}$, along the channel axes, which ensures inter-spatial relationship of features. 

\subsection{Profile to Frontal Coupled Subspace Learning Network}

Our goal is to learn a discriminative, pose-invariant feature representation from a pair of face images to a compact embedding subspace such that we can perform matching of profile face images with respect to a gallery of frontal face images in the embedded domain. Therefore, to learn a rich feature representation, we propose a coupled deep convolutional network guided by a pose attention module. Our method adopts a variant of InceptionResnet \cite{szegedy2017inception} architecture that is used in FaceNet \cite{schroff2015facenet} as the core element of our network. It is pretrained on VGGFace2\cite{cao2018vggface2}. 
 
 The model structure in Figure 2 illustrates that a pair of images goes through the coupled network consisting of two dedicated branches to extract features from both frontal and profile images. Since there exists pose variations in profile faces, we hypothesize that we can leverage pose as an auxiliary information to improve the ability of extracting highly discriminative features from these profile faces. To accomplish this, the profile image is also fed to a pretrained Hopenet pose estimation network. It provides a pose attended information via our PAB module to sequentially distills features along both channel and spatial dimension of our profile encoder.

Previous section explains the block design of our PAB module. It consists of two sub-modules: (1) ACAM, that generates a 1D channel attention map of $1,792$ to refine the feature maps ($1,792\times3\times3$) of our profile coupled network along the channel dimension, and (2) SpAM, which produces a 2D spatial attention map of size 3x3 to spatially attend the informative region in the feature maps of our profile network as in Figure 2. For more details note that, both 1D and 2D attention maps are multiplied with the feature maps of the profile network for adaptive feature refinement.

Such sharing of pose as an auxiliary information during feature extraction from profile faces results in informative and task relevant features, which otherwise would not have been attained from training only with a massive labelled faces. In addition, it also allows for better generalization of the PIFR task by looking at new interpretations of the features. Once the embeddings are established as feature vectors, we optimize the network via class-specific contrastive loss. It tries to minimize squared Euclidean distance between the features of positive pairs (i.e., when profile and frontal image share the same identity) and
maximize it for negative pairs (i.e., when profile and frontal image comes from different identities).
\vspace{-2.8mm}
\section{Loss Function}
\vspace{-0.2cm}
Our goal is to learn a compact 512-D embedding subspace by coupling two mapping networks, one for frontal and another one for profile face image, via contrastive loss, $L_{cont}$ \cite{chopra2005learning}.
We compute this loss metric, $L_{cont}$ over a set of genuine (i.e., a profile face image of a subject with its corresponding frontal face image) and imposter (i.e., a profile face image of a subject and a frontal face image of a different subject) pairs such that images belonging to the same identity (genuine pair) are embedded as close as possible. Simultaneously, images of different identities are pushed away from each other in the common embedded subspace. The contrastive loss function is formulated as:

 \begin{equation}
\begin{split}
L_{cont}(z&(x^i_{p}),z(x^j_{f}),Y)= \\ & 
 (1-Y)\frac{1}{2}(D_z)^2 + (Y)\frac{1}{2}(\mbox{max}(0,m-D_z))^2,  
 \end{split}
 \end{equation}where
$x^i_{p}$ and $x^j_{f}$ denote the input profile and frontal face images, respectively. The variable $Y$ is a binary label, which is equal to 0 if $x^i_{p}$ and $x^j_{f}$ belong to the same class (i.e., genuine pair), and equal to 1 if $x^i_{p}$ and $x^j_{f}$ belong to the different class (i.e., impostor pair). $z(.)$ is used to denote the mapping function for $x^i_{p}$ and $x^j_{f}$ into a compact embedding subspace. To \enquote{tighten} the constraint, $m$ is used as contrastive margin.
 
The Euclidean distance, $D_z$, between the embedding features, $z(x^i_{p})$ and $z(x^j_{f})$, is given by:

 \begin{equation}
     D_z=\left\lVert z(x^i_{p})-z(x^j_{f})\right\rVert_2.
 \end{equation}
Therefore, if $Y=0$ (i.e., genuine pair), then the contrastive loss function $(L_{cont})$ is given as:
 \begin{equation}
L_{cont}(z(x^i_{p}),z(x^j_{f}),Y)  = \frac{1}{2}\left\lVert z(x^i_{p})-z(x^j_{f})\right\rVert^2_2, 
\end{equation}

and if $Y=1$ (i.e., impostor pair), then contrastive loss function $(L_{cont})$ is :
  \begin{equation}
  \begin{split}
L_{cont}(z(x^i_{p}),&z(x^j_{f}),Y)  = \\ & \frac{1}{2}\mbox{max}\biggl(0,m-\left\lVert z(x^i_{p})-z(x^j_{f})\right\rVert^2_2\biggr).
\end{split}
\end{equation}

Thus, the total loss to optimize the entire network is denoted by $L_{total}$ for coupling both the profile and frontal face in the embedded domain: 

\vspace{-0.25cm}
\begin{equation}
    \begin{split}
        L_{total}=\frac{1}{N^2}\sum_{i=1}^{N}\sum_{j=1}^{N}L_{cont}(z(x^i_{p}),z(x^j_{f}),Y), 
    \end{split}\label{eq:6}
\end{equation}
where \textit{N} is the number of training samples. The main purpose of using the contrastive loss is to be able to use the class labels, and margin to ensure discriminative embedding subspace, which may not be obtained with some other metric such as the Euclidean distance. Finally, we use this pose attended discriminative embedding subspace for matching of the profile images with the frontal images.


\section{Experiments}
In this section, we describe our implementation details and the datasets that we have used to conduct our experiments. To  evaluate the performance of our proposed PIFR network, we experiment under two settings: (1) face identification on controlled Multi-PIE\cite{gross2010multi} face dataset, and (2) face verification/identification on in-the-wild datasets including CFP\cite{sengupta2016frontal} and IJB-C\cite{maze2018iarpa} with their official evaluation protocols. In addition, we also report face recognition accuracy compared to several state-of-the-art results on these datasets.  

\subsection{Datasets}

\textbf{Multi-PIE}: The  Multi-PIE dataset is the largest dataset released for multi-view face recognition with respect to controlled variations in illumination and expressions across different poses. It contains 7,54,204 images of 337 identities, captured at 15 view points ranging from $-90\degree\sim+90\degree$, over 20 illumination conditions. To evaluate our proposed method for identification task, we conduct experiments under two settings following protocol used in\cite{zhao2018towards} for fair comparison. \textbf{Setting 1} includes images only from session 1 in the Multi-PIE dataset, which has 250 subjects. For training, we choose first 150 identities with 11 poses within $\pm 90\degree$ and 20 illuminations. During testing, one frontal view with neutral expression and illumination (i.e., ID 07) is used in the gallery for each of the remaining 100 identities and other images are considered as probe images. \textbf{Setting 2} includes images only with neutral expression over all four sessions providing 337 identities. To train our network, we use first 200 identities, while rest of the 137 IDs have been used for testing. We maintain similar setup as setting 1 for our gallery and probe. 
\vspace{0.1cm}
\textbf{CFP}: The Celebrities in Frontal-Profile (CFP) dataset is introduced to handle large-pose variations. It contains identities of 500 celebrities, which have been collected under constrained (i.e., images at different pose, illumination and expression) and unconstrained (i.e., images collected from the Internet) settings. For each celebrity, it includes 10 frontal and 4 profile images. Following their standard 10-fold evaluation protocol\cite{sengupta2016frontal}, we split the dataset into 10 folds, each with 350 genuine and 350 impostor pairs to perform both frontal-to-frontal (FF), and frontal-to-profile (FP) verification task.
\vspace{0.1cm}
\textbf{IJB-C}: The IARPA Janus Benchmark–C (IJB-C)\cite{maze2018iarpa} face dataset has been released to advance the unconstrained face recognition by modeling more practical face recognition use cases. It is an extension to the publicly available IJB-B\cite{whitelam2017iarpa} dataset, which contains $3,531$ subjects with extreme variations in expression, illumination, geographic origin, and more. In total, it has $31,334$ still images and $1,17,542$ video frames collected in unconstrained settings with different protocols. To evaluate our algorithm's ability, we perform both face verification (1:1), and identification (1:N) tasks following their protocol.

\subsection{Implementation Details}
\noindent To implement our proposed coupled learning framework, we have used InceptionResnet-v1 \cite{szegedy2017inception} pretrained on VGG-Face2 dataset. Since it is difficult to train the entire network from scratch, we freeze all the trained layers before average pooling for both frontal and profile mapping modules as shown in Figure 2. At the same time, our PAB module provides a pose attended 1D channel attention map, and a 2D spatial attention map to assist the profile network to use only the relevant features while extracting deep features from the profile faces. Therefore, the gradient also flows back through this PAB module to update its weights during optimization. Note that, since misleading pose information can misguide the training, we don't train Hopenet, which has been already trained on a very large dataset, and proved to be an efficient pose estimation model. The entire framework has been implemented in Pytorch. We used a batch size of 32 and the Adam optimizer \cite{Kingma2015AdamAM} with first-order momentum of 0.5, and learning rate of $10^{-3}$. For training, we generate same number of genuine, and impostor pairs from frontal, and profile images of the same/different subjects to avoid biasness towards positive pairs. 

\vspace{-2mm}
\begin{table}[t]
\centering

\caption{Rank-1 recognition rates ($\%$) across poses and illuminations under Multi-PIE Setting-1.}
\scalebox{0.63}{\begin{tabular}{c c c c c c c}
 \hline
\multicolumn{1}{c}{Method} &\multicolumn{1}{c}{$\pm90\degree$} &\multicolumn{1}{c}{$\pm75\degree$}&\multicolumn{1}{c}{$\pm60\degree$}&\multicolumn{1}{c}{$\pm45\degree$}&\multicolumn{1}{c}{$\pm30\degree$}&\multicolumn{1}{c}{$\pm15\degree$}\\ [0.5ex] 
 \hline \hline
HPN \cite{ding2017pose} & $29.8$&$47.5$&$61.2$&$72.7$&$78.2$&$84.2$\\
c-CNN \cite{xiong2015conditional} & $47.2$&$60.7$&$74.4$&$89.0$&$94.1$&$97.0$\\
TP-GAN \cite{huang2017beyond} & $64.0$&$84.1$&$92.9$&$98.6$&$99.99$&$99.8$\\
PIM \cite{zhao2018towards} & $75.0$&$91.2$&$97.7$&$98.3$&$99.4$&$99.8$\\
CAPG-GAN \cite{hu2018pose} & $77.1$&$87.4$&$93.7$&$98.3$&$99.4$&$99.9$\\
FNM$+$VGG-Face \cite{qian2019unsupervised} & $41.1$&$67.3$&$83.6$&$93.6$&$97.2$&$99.0$\\
FNM$+$Light CNN \cite{qian2019unsupervised}& $55.8$&$81.3$&$93.7$&$98.2$&$99.5$&$99.9$\\
PF-cpGAN \cite{taherkhani2020pf}& $88.1$&$94.2$&$97.6$&$98.9$&$99.9$&$99.9$\\
Backbone(without attention)& $75.68$&$98.20$&$\textbf{100.0}$&$\textbf{100.0}$&$\textbf{100.0}$&$\textbf{100.0}$\\
\textbf{Ours}  & $\textbf{89.5}$&$\textbf{98.7}$&$\textbf{100.0}$&$\textbf{100.0}$&$\textbf{100.0}$&$\textbf{100.0}$\\
\hline
\vspace{-11.5mm}

\end{tabular}}
\end{table}
\subsection{Evaluations on the Multi-PIE Benchmark}
\vspace{-2mm}
To show the effectiveness of our proposed method, first we evaluate our model on a controlled database, Multi-PIE for profile-to-frontal face recognition task under two different settings. We compare our method with several state-of-the-art PIFR algorithms including HPN \cite{ding2017pose}, c-CNN \cite{xiong2015conditional}, PIM \cite{zhao2018towards}, FNM \cite{zhao2018towards}, and competitive GAN-based methods : TP-GAN \cite{huang2017beyond}, CAPG-GAN \cite{hu2018pose}, and PF-cpGAN \cite{taherkhani2020pf}. 

Table 1 shows our rank-1 recognition accuracy compared to other approaches across full yaw variations and illuminations under setting-1. For this experimental setup, we consistently achieve 100\% accuracy over yaw angles $<$ $75\degree$, while outperforming other baselines. Even under extreme pose (i.e., $\pm75\degree$, and $\pm90\degree$), when compared to CAPG-GAN, and PF-cpGAN, we significantly outperform them by achieving average 11.85\%, and 3\% higher accuracy, respectively. Compared to the performance of our backbone coupled network (without attention), we achieve \textbf{13.82\%} more recognition accuracy for full profile face ($\pm90\degree$).
\vspace{2mm}
We also assess the performance of our proposed network on faces in Multi-PIE under setting-2, which consists of more challenging face identities than setting-1. Evaluation results, shown in Table 2 suggests that our proposed PAB module does assist the face recognition network to achieve 2.3\%, and 2.7\% increase over the best performing method, PIM \cite{zhao2018towards} in the large pose variations; $\pm90\degree$, and $\pm75\degree$, respectively. Apart from this, our network achieves superior performance over the other baseline models \cite{huang2017beyond, yin2017towards, zhao2018towards} in all yaw angles. Similar to setting 1, we note \textbf{14.0\%} improvement on backbone (without attention) for $\pm90\degree$ profile faces as well. These improvements indicate the efficacy of our method for PIFR in a constrained environment.

\begin{table}[t]
\centering
\caption{Rank-1 recognition rates ($\%$) across poses and illuminations under Multi-PIE Setting-2.}
\scalebox{0.63}{\begin{tabular}{c c c c c c c}
 \hline
\multicolumn{1}{c}{Method} &\multicolumn{1}{c}{$\pm90\degree$} &\multicolumn{1}{c}{$\pm75\degree$}&\multicolumn{1}{c}{$\pm60\degree$}&\multicolumn{1}{c}{$\pm45\degree$}&\multicolumn{1}{c}{$\pm30\degree$}&\multicolumn{1}{c}{$\pm15\degree$}\\ [0.5ex] 
 \hline \hline
FF-GAN \cite{yin2017towards} & $61.2$&$77.2$&$85.2$&$89.7$&$92.5$&$94.6$\\
TP-GAN \cite{huang2017beyond} & $64.6$&$77.4$&$87.7$&$95.4$&$98.0$&$98.6$\\
CAPG-GAN \cite{hu2018pose} & $66.0$&$83.05$&$90.6$&$97.3$&$99.5$&$99.8$\\
DA-GAN \cite{yin2020dual} & $81.5$&$93.2$&$97.2$&$99.1$&$99.8$&$99.9$\\
PIM \cite{zhao2018towards} & $86.5$&$95.0$&$98.1$&$98.3$&$98.5$&$99.0$\\
Backbone (without attention)  & $74.8$&$96.8$&$\textbf{100.0}$&$\textbf{100.0}$&$\textbf{100.0}$&$\textbf{100.0}$\\
\textbf{Ours}& $\textbf{88.8}$&$\textbf{97.7}$&$\textbf{100.0}$&$\textbf{100.0}$&$\textbf{100.0}$&$\textbf{100.0}$\\

\hline

\end{tabular}}
\label{table:multi-pie-yaw}
\end{table}

\begin{table}[t]
\centering

\caption{Performance comparison on CFP dataset. Mean Accuracy and equal error rate (EER) with standard deviation over 10 folds.}
\scalebox{0.60}{\begin{tabular}{c c c c c c}
 \hline
 &\multicolumn{2}{c}{Frontal-Profile (FP)} &\multicolumn{2}{c}{Frontal-Frontal (FF)}\\ [0.5ex] 
 \hline
 Algorithm&Accuracy&EER&Accuracy&EER  \\ \hline
FV+DML \cite{sengupta2016frontal}  &58.47(3.51) &38.54(1.59)&91.18(1.34)&8.62(1.19) \\ \hline
LBP+Sub-SML \cite{sengupta2016frontal} &70.02(2.14)&29.60(2.11)&77.98(1.86)&16.00(1.74)\\ \hline
HoG+Sub-SML \cite{sengupta2016frontal} &77.31(1.61)&22.20(1.18)&85.97(1.03)&11.45(1.35) \\ \hline
FV+Sub-SML \cite{sengupta2016frontal} &80.63(2.12)&19.28(1.60)&88.53(1.58)&8.85(0.74)\\ \hline
Deep Features \cite{sengupta2016frontal} &84.91(1.82)&14.97(1.98)&93.00(1.55)&3.48(0.67) \\ \hline
Triplet Embedding \cite{sankaranarayanan2016triplet} &89.17(2.35)&8.85(0.99)&98.88(1.56)&2.51(0.81) \\ \hline
Light CNN-29 \cite{wu2018light} &92.47(1.44)& 8.71(1.80)& 99.64(0.32)& 0.57(0.40)\\ \hline
PIM (Light CNN-29) \cite{wu2018light} &93.10(1.01)&7.69(1.29)&99.44(0.36)& 0.86(0.49) \\ \hline
PR-REM \cite{cao2018pose} &93.25(2.23)&7.92(0.98)&98.10(2.19)&1.10(0.22) \\ \hline
PF-cpGAN \cite{taherkhani2020pf} &93.78(2.46)&7.21(0.65)&98.88(1.56)&0.93(0.14) \\ \hline
Backbone (without attention) &92.57(1.10)& 4.24(0.54)& 97.10(0.11)&1.5(0.25) \\ \hline
\textbf{Ours} &\textbf{95.67(1.64)}&\textbf{2.02}(0.62)&\textbf{99.70(0.21)}&\textbf{0.55(0.35)} \\ \hline
\textcolor{blue}{Human}\cite{sengupta2016frontal} & \textcolor{blue}{94.57(1.10)}& \textcolor{blue}{5.02(1.07)}& \textcolor{blue}{96.24(0.67)}& \textcolor{blue}{5.34(1.79)}\\ \hline
 
\vspace{-9mm}
\end{tabular}}
\label{table:table_cfp}
\end{table}

\subsection{Evaluations on the CFP Benchmark}

We evaluate our proposed method on the Celebrities in Frontal-Profile (CFP) dataset to analysis face verification in unconstrained environment. To perform evaluation, we follow the standard 10-fold protocol like other approaches in the literature. We report the mean and standard deviation of accuracy, and Equal Error Rate (EER) over the 10 splits for both frontal-frontal (FF) and frontal-profile (FP) face verification settings.

Table 3 shows a comparison of our method with other state-of-the-art face recognition performance on the CFP benchmark datatset. For fair comparison, first, we consider three different hand-crafted feature extraction techniques: Hog\cite{dalal2005histograms}, LBP\cite{ahonen2006face}, and Fisher Vector\cite{simonyan2013fisher} along with metric learning techniques Sub-SML\cite{cao2013similarity}, and diagonal metric learning (DML)\cite{cao2013similarity}. To compare against deep learning based approaches, we include Deep Features\cite{chen2016unconstrained}, Triplet Embedding\cite{sankaranarayanan2016triplet}, Light CNN-29\cite{wu2018light}, and recently proposed GAN-based latent feature learning framework, PF-cpGAN\cite{taherkhani2020pf}. 

From the results summarized in Table 3, we observe that our proposed method outperforms human performance for both FF and FP settings. It also makes substantial improvement over the conventional hand-crafted features by achieving average 18\% higher accuracy with 24\% decrease in EER for more challenging FP setting. In addition, when compared to best performing PF-cpGAN, our proposed method improves the accuracy by 1.89\% and reduce EER significantly by 5\% for FP verification. We also improve on the performance of PR-REM\cite{cao2018pose} by 2.5\% higher accuracy with approximately 6\% lower EER. Moreover, we obtain 3.10\% better accuracy than the backbone network where we do not apply any attention.

\vspace{-0.2cm} 

\begin{table}[t]
\centering

\caption{Performance evaluation on IJB-C benchmark. 
Symbol ’-’ indicates that the metric is not available for that protocol.}
\Large
\scalebox{0.37}{\begin{tabular}{c c c c c}
 \hline
\multicolumn{1}{c}{\multirow{2}{*}{Method}} &\multicolumn{2}{c}{1:1 Verification} &\multicolumn{2}{c}{1:N Identification}\\ [0.5ex] 
 \cline{2-5}
 &GAR@ FAR$=0.01$&GAR@ FAR$=0.001$&@ Rank-1 &@ Rank-5  \\ \hline \hline
 GOTS \cite{maze2018iarpa} &$61.99$&$33.4$&$38.5$&$53.8$ \\
 FaceNet \cite{schroff2015facenet} &$ 81.76$& $66.45$ &$69.22$&$78.7$ \\
 VGGFace \cite{parkhi2015deep}
 &$87.13$&$74.79$&$78.60$&$87.2$ \\
 CFR-GAN \cite{ju2022complete} &86.46&74.81&-&-\\
 FNM\cite{qian2019unsupervised} &91.2 &80.4&78.6&88.7\\
 PR-REM\cite{cao2018pose}&90.6 &80.2 &77.1 &87.6\\
Backbone(without attention)&$89.1$&$79.9$&$71.8$&$81.2$\\
\textbf{Ours}&$\textbf{92.8}$&$\textbf{82.5}$&$\textbf{80.33}$&$\textbf{90.42}$ \\ \hline
 \vspace{-16.5mm}
\end{tabular}}
\label{table:table_ijb}
\end{table}

\subsection{Evaluations on the IJB-C Benchmark}

We further evaluate face recognition (i.e., verification and identification) on another challenging benchmark IJB-C,  to validate the superiority of our proposed method in unconstrained environment. We compare with the recent state-of-the-art algorithms CFR-GAN \cite{ju2022complete}, FNM \cite{qian2019unsupervised}, and PR-REM \cite{cao2018pose}, along with prior works \cite{parkhi2015deep, schroff2015facenet} in \cite{maze2018iarpa} for fair evaluation. As shown in Table 4, for profile to frontal verification, we improve the genuine accept rate (GAR) by approximately $7.69\%$, and $2.1\%$ at the false accept rate (FAR) of 0.001 compared to recent works\cite{ju2022complete, qian2019unsupervised}. Moreover, we also obtain outstanding performances on identification. Specifically, we achieve $1.73\%$, and $3.23\%$ higher recognition accuracy for rank-1 in comparison to the  FNM, and PR-REM, respectively.
To show the significant contribution of our proposed PAB module, we compare our results with the backbone network (without attention). It shows our idea of incorporating pose information boosts FP verification performance by $2.60\%$ at 0.001 FAR, and identification accuracy by $8.53\%$ for rank-1.

\subsection{Frontal Face Reconstruction from Pose-Invariant Features Learned in Deep Subspace}

The purpose of our proposed PAB is to enhance the recognition performance of our coupled deep subspace learning framework via contributing in feature refinement. In addition, class-specific contrastive loss has been used to push the network achieve pose-invariance in the embedding feature domain. To validate our hypothesis,  previous sections show comprehensive analysis on verification and identification task for both constrained and in the wild conditions. Apart from recognition task, there are many other scopes to utilize the feature vector learned in the deep subspace. For instance, if we could reconstruct frontal face from the deep features of its corresponding profile face, it can be used in many face analysis tasks including emotion detection, expression tracking etc. Moreover, it has broad applications in vision, graphics, and robotics. Therefore, to demonstrate the usefulness of our proposed method, we adopt a GAN model\cite{tian2018cr} for reconstruction.
\begin{figure}[t]
\begin{center}
\includegraphics[width=.60\linewidth]{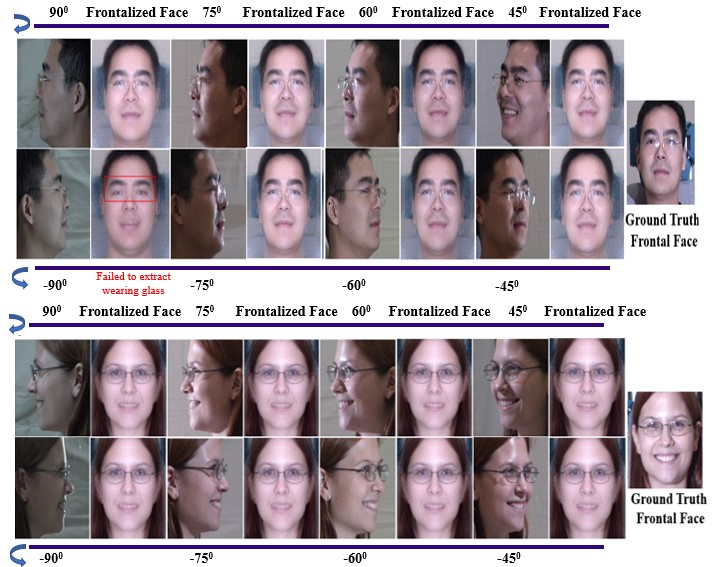}
\end{center}
   \vspace*{-3mm}
   \caption{Reconstruction results via a GAN\cite{tian2018cr} model on Multi-PIE across different pose, illumination and expression using the compact 512-D embedding features learned using our proposed network. }

\vspace{-5mm}
\end{figure}
To accomplish this task, we modify residual network used in the two-pathway encode-decoder architecture proposed by Tian et al. \cite{tian2018cr}. We consider their decoder module with the discriminator network for frontal face synthesis. For training/testing, we select profile and corresponding frontal images from setting 1 of Multi-PIE dataset. First, we extract 512-D embedding feature vector from our proposed PAB guided pose-invariant face recognition network for each of the profile image in the trainset. After that, these profile feature vectors are given as input to the decoder and corresponding frontal faces with no expression and neutral illumination are used as target, which force the network adversarially learn the image distribution of the frontal faces. To generate identity preserving, high visual quality frontal faces from its profile deep features, we incorporate pixel-wise $L_{1}$ reconstruction error, VGG-16 based Perceptual loss \cite{johnson2016perceptual}, and Light CNN-29 \cite{wu2018light} network for identification loss along with adversarial loss. In Figure 4, we show some representative results on Multi-PIE test samples. Reconstruction results indicate that our proposed pose attention-guided coupled framework is able to provide robust, and discriminative features in the deep subspace for multiple use of profile to frontal matching in the embedded domain as well as high-fidelity frontal face synthesis. 

\vspace{-2mm}
\section{Ablation Study}
\subsection{Embedding Dimensionality}
\begin{table}
\centering
\vspace{1mm}
\caption{Rank-1 recognition rates (\%) on Multi-PIE (Setting-1) for different embedding dimensions. We select 512-D for all experiments reported in this paper.}
\scalebox{0.60}{\begin{tabular}{c c c}
 \hline
\multicolumn{1}{c}{Dims}&\multicolumn{1}{c}{$\pm90\degree$} &\multicolumn{1}{c}{$\pm75\degree$}\\ [0.5ex] 
 \hline  \hline
128 & $60.8$&$67.5$\\
256 & $70.2$&$78.1$\\
512  & $\textbf{89.5}$&$\textbf{98.7}$\\
\hline
\vspace{-9mm}
\end{tabular}}
\end{table}

To represent each face into a tightly compact embedding subspace, we explore different embedding dimensionalities: 128, 256, and 512. Experimental results reported in Table 5 illustrates that the network is able to extract features enriched with relevant information in 512 dimension.  
\vspace{-0.1cm}
\subsection{Attention Map}
In this section, we show the effective design approach of our proposed pose attention mechanism to efficiently guide the face recognition network. We first focus on computing different approaches of SpAM in the pose attention block, PAB. Finally, we observe different combination of channel and spatial attention in deep profile feature extraction. Each experiment has been explained in the following sections.
\vspace{-0.3cm}
\subsubsection{Spatial Attention} Given the channel-wise refined features, we explore two different approaches to generate a 2D spatial attention map: (1) first, we use average-and max-pooling across the channel axes, to generate two 2D descriptors, and then apply standard 1 × 1 convolution followed by a max pool layer. (2) second, we similarly generate two 2D descriptors, and apply $3\times3$ convolution with stride 2, which proves to be outperforming the first approach. We report the comparison of two methods in Table 6.
 
\begin{table}[t]
\centering
\caption{Rank-1 recognition rates (\%) on Multi-PIE (Setting-1) for different approaches of spatial attention}
\scalebox{0.65}{\begin{tabular}{c c c}
 \hline
\multicolumn{1}{c}{Description}&\multicolumn{1}{c}{$\pm90\degree$} &\multicolumn{1}{c}{$\pm75\degree$}\\ [0.5ex] 
 \hline
Channel Refined Features + $1\times1$ conv + Max Pool& $87.8$&$94.5$\\
Channel Refined Features +  $3\times3$ conv + Stride-2 & $\textbf{89.5}$&$\textbf{98.7}$\\
\hline
\end{tabular}}
\end{table}
\vspace{-4mm}

\subsubsection{Arrangement of Spatial and Channel Attention}
In this experiment, we apply channel and spatial attention in two different ways. From a spatial viewpoint, the channel attention works to attend global information whereas the spatial attention focuses on local neighbourhood. However, the network response can be different upon the sequential order of each attention mechanism. Table 7 summarizes the recognition performance on Multi-PIE for different attention sequences. The results show that the we achieve better performance when we use channel-spatial order rather than the vise versa. 

\begin{table}[t]
\centering
\caption{Rank-1 recognition rates (\%) on Multi-PIE (Setting-1) for different arrangements in attention mechanism}
\scalebox{0.65}{\begin{tabular}{c c c}
 \hline
\multicolumn{1}{c}{Description}&\multicolumn{1}{c}{$\pm90\degree$} &\multicolumn{1}{c}{$\pm75\degree$}\\ [0.5ex] 
 \hline

InceptionResnet + spatial + channel  & $87.2$&$95.6$\\
InceptionResnet + channel + spatial& $\textbf{89.5}$&$\textbf{98.7}$\\
\hline
\vspace{-9mm}
\end{tabular}}
\end{table} 
\vspace{-0.2cm}

\subsection{Visualization}

\begin{figure}[t]
\begin{center}
\includegraphics[width=0.8\linewidth]{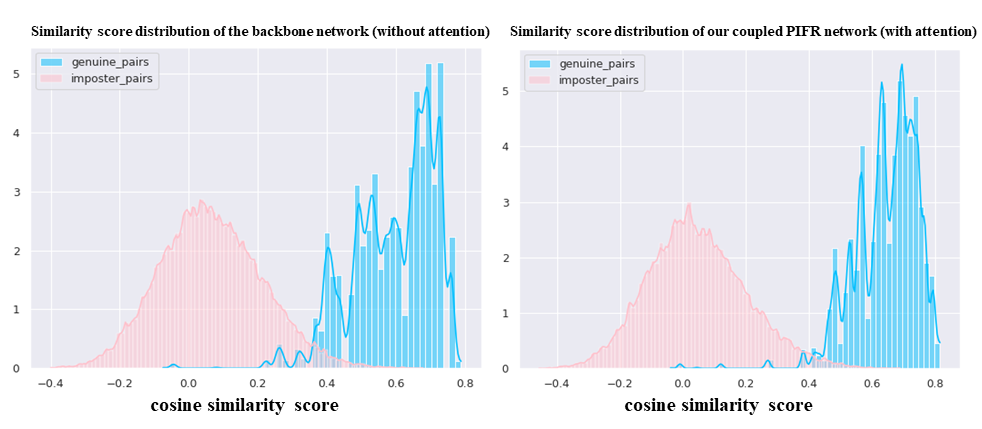}
\end{center}
  \vspace{-5mm}
   \caption{Comparing Cosine similarity distributions of the genuine pairs and imposter pairs for full profile faces ($\pm90\degree$) of Multi-Pie Setting 1 between the backbone network (without attention) and our coupled PIFR network (with attention)}

\vspace{-0.7cm}
\end{figure}

As shown in Figure 5, when compared
to the backbone network (without attention), the similarity distributions of the genuine pairs and the imposter pairs in our proposed coupled PIFR network are more compact and distinct for full profile variations ($\pm90\degree$). Moreover, the area of similarity between genuine pairs spread more and overlap with the area of imposter pairs when we only train backbone network with constrastive loss without imposing attention on it. It further supports our proposed idea of pose refinement via PAB attention module.
\vspace{-1mm}
\section{Conclusion}
In this paper, we propose a novel perspective of leveraging pose as auxiliary information to guide a coupled profile to frontal deep subspace learning framework for PIFR. A PAB module is designed to distill pose-specific useful features from profile faces in deep convolutional layers. To ensure discriminative, pose-invariant feature representation into a compact embedding subspace, we couple both profile and frontal face images via a contrastive loss, which maximizes the pair-wise similarity in the embedded domain. We perform a comprehensive experiments on several benchmark datasets both in controlled and uncontrolled environmental settings to evaluate the robustness of our model. The results indicate that our model remarkably outperform other state-of-the-art algorithms for profile-to-frontal pose-invariant face recognition. In addition, we conduct a quick experiment to explore the generative capability of the embedding features learned in deep subspace of our network. Moreover, we also investigate embedding dimensionality and attention mechanisms from different perspectives to offer an effective design choice of our proposed network. 
\vspace{-2mm}
\section{Acknowledgements}

This research is based upon work supported by the Office of the Director of National Intelligence (ODNI), Intelligence Advanced Research Projects Activity (IARPA), via IARPA R\&D Contract No. 2022-21102100001. The views and conclusions contained herein are those of the authors and should not be interpreted as necessarily representing the official policies or endorsements, either expressed or implied, of the ODNI, IARPA, or the U.S. Government. The U.S. Government is authorized to reproduce and distribute reprints for Governmental purposes notwithstanding any copyright annotation thereon.

{\small
\bibliographystyle{ieee}
\bibliography{egbib}
}

\end{document}